\pdfoutput=1
\documentclass[runningheads]{llncs}

\usepackage{graphicx}
\usepackage{amsmath,amssymb}
\usepackage{hyperref}

\usepackage{tikz}
\usetikzlibrary{arrows,arrows.meta,backgrounds,calc,patterns,positioning,shapes,shadows}
\usepackage[linesnumbered, ruled, noend]{algorithm2e}
\usepackage[capitalise]{cleveref}

\usepackage{csquotes}
\usepackage{subcaption}
\usepackage[nolist, nohyperlinks]{acronym}
\acrodef{cp}[CP]{colour passing}
\acrodef{fg}[FG]{factor graph}
\acrodef{kl}[KL]{Kullback-Leibler}
\acrodef{lifg}[LIFAGU]{Lifting Factor Graphs with Some Unknown Factors}
\acrodef{ljt}[LJT]{lifted junction tree}
\acrodef{lv}[logvar]{logical variable}
\acrodef{lve}[LVE]{lifted variable elimination}
\acrodef{pf}[parfactor]{parametric factor}
\acrodef{pfg}[PFG]{parameterised factor graph}
\acrodef{prv}[PRV]{parameterised randvar}
\acrodef{rv}[randvar]{random variable}
\acrodef{ve}[VE]{variable elimination}
\acrodef{wl}[WL]{Weisfeiler-Leman}

\newcommand{\abs}[1]{\lvert #1 \rvert}
\newcommand{\Ne}{\ensuremath{\mathrm{Ne}}}
\newcommand{\twostep}[1]{\ensuremath{2\text{-}step(#1)}}
\newcommand{\twostepG}[2]{\ensuremath{2\text{-}step_{#1}(#2)}}

\definecolor{myyellow}{RGB}{247,192,26}
\definecolor{myblue}{RGB}{37,122,164}
\definecolor{mygreen}{RGB}{78,155,133}
\definecolor{mypurple}{RGB}{86,51,94}

\definecolor{newblue}{RGB}{50,113,173}
\definecolor{newred}{RGB}{222,32,36}
\definecolor{newgreen}{RGB}{70,165,69}
\definecolor{newpurple}{RGB}{140,69,152}

\pgfdeclarelayer{bg}
\pgfsetlayers{bg,main}

\tikzset{
	rv/.style={draw, ellipse},
	pf/.style={draw, rectangle, draw = black!70, fill = gray!30},
	arc/.style = {->, semithick, >={[round,sep]Stealth}},
}

\newcommand\factor[6]{
	\node[pf, #1=#3 of #2, label=#4:{#5}](#6) {};
}

\newcommand\nodecolorshift[5]{
	\node[circle, fill=#1, above right=0.1cm of #2, inner sep=0pt, minimum size=2mm, xshift=#4, yshift=#5](#3) {};
}

\newcommand\factorcolor[3]{
	\node[circle, fill=#1, above right=0cm and 0.05cm of #2, inner sep=0pt, minimum size=2mm](#3) {};
}
\newcommand\factorcolorshift[4]{
	\node[circle, fill=#1, above right=0cm and 0.05cm of #2, inner sep=0pt, minimum size=2mm, xshift=#4](#3) {};
}

\newcommand\pfs[8]{
	\node[pf, #1=#3 of #2, xshift=-1mm, yshift=1mm](#6) {};
	\node[pf, #1=#3 of #2, label={[label distance=1mm]#4:{#5}}](#7) {};
	\node[pf, #1=#3 of #2, xshift=1mm, yshift=-1mm](#8) {};
}

\begin{document}
\title{Lifting Factor Graphs with Some Unknown Factors}

\author{
	Malte Luttermann\inst{1,2}\orcidID{0009-0005-8591-6839} \and
	Ralf Möller\inst{1,2}\orcidID{0000-0002-1174-3323} \and
	Marcel Gehrke\inst{1}\orcidID{0000-0001-9056-7673}
}
\institute{
	Institute of Information Systems, University of Lübeck, Germany \\
	\and
	German Research Center for Artificial Intelligence (DFKI), Lübeck, Germany \\
	\email{\{luttermann,moeller,gehrke\}@ifis.uni-luebeck.de}
}
\maketitle

\begin{abstract}
	Lifting exploits symmetries in probabilistic graphical models by using a representative for indistinguishable objects, allowing to carry out query answering more efficiently while maintaining exact answers.
	In this paper, we investigate how lifting enables us to perform probabilistic inference for factor graphs containing factors whose potentials are unknown.
	We introduce the \emph{\ac{lifg} algorithm} to identify symmetric subgraphs in a factor graph containing unknown factors, thereby enabling the transfer of known potentials to unknown potentials to ensure a well-defined semantics and allow for (lifted) probabilistic inference.
\end{abstract}
\acresetall

\section{Introduction}
To perform inference in a probabilistic graphical model, all potentials of every factor are required to be known to ensure a well-defined semantics of the model.
However, in practice, scenarios arise in which not all factors are known.
For example, consider a database of a hospital containing patient data and assume a new patient arrives and we want to include them into an existing probabilistic graphical model such as a \ac{fg}.
Clearly, not all attributes of the database are measured for every new patient, i.e., there are some values missing, resulting in an \ac{fg} with unknown factors and ill-defined semantics when including a new patient in an existing \ac{fg}.
Therefore, we aim to add new patients to an existing group of indistinguishable patients to treat them equally in the \ac{fg}, thereby allowing for the imputation of missing values under the assumption that there exists such a group for which all values are known.
In particular, we study the problem of constructing a lifted representation having well-defined semantics for an \ac{fg} containing unknown factors---that is, factors whose mappings from input to output are unknown.
In probabilistic inference, lifting exploits symmetries in a probabilistic graphical model, allowing to carry out query answering more efficiently while maintaining exact answers~\cite{Niepert2014a}.
The main idea is to use a representative of indistinguishable individuals for computations.
By lifting the probabilistic graphical model, we ensure a well-defined semantics of the model and allow for tractable probabilistic inference with respect to domain sizes.

Previous work on constructing a lifted representation builds on the \acl{wl} algorithm~\cite{Weisfeiler1968a} which incorporates a colour passing procedure to detect symmetries in a graph, e.g. to test for graph isomorphism.
To construct a lifted representation for a given \ac{fg} where all factors are known, the \ac{cp} algorithm (originally named \enquote{CompressFactorGraph})~\cite{Ahmadi2013a,Kersting2009a} is commonly used.
Having obtained a lifted representation, algorithms performing lifted inference can be applied.
A widely used algorithm for lifted inference is the \acl{lve} algorithm, first introduced by Poole~\cite{Poole2003a} and afterwards refined by many researchers to reach its current form~\cite{DeSalvoBraz2005a,DeSalvoBraz2006a,Kisynski2009a,Milch2008a,Taghipour2013a}.
Another prominent algorithm for lifted inference is the \acl{ljt} algorithm~\cite{Braun2016a}, which is designed to handle sets of queries instead of single queries.

To encounter the problem of constructing a lifted representation for an \ac{fg} containing unknown factors, we introduce the \ac{lifg} algorithm, which is a generalisation of the \ac{cp} algorithm.
\Ac{lifg} is able to handle arbitrary \acp{fg}, regardless of whether all factors are known or not. By detecting symmetries in an \ac{fg} containing unknown factors, \ac{lifg} generates the possibility to transfer the potentials of known factors to unknown factors to eliminate unknown factors from an \ac{fg}.
We show that, under the assumption that for every unknown factor there is at least one known factor having a symmetric surrounding graph structure to it, \emph{all} unknown potentials in an \ac{fg} can be replaced by known potentials.
Thereby, \ac{lifg} ensures a well-defined semantics of the model and allows for lifted probabilistic inference.

The remaining part of this paper is structured as follows.
\Cref{sec:prelim} introduces necessary background information and notations. We first briefly recapitulate \acp{fg}, afterwards define \acp{pfg}, and then describe the \ac{cp} algorithm as a foundation for \ac{lifg}.
Afterwards, in \cref{sec:lifg}, we introduce \ac{lifg} as an algorithm to obtain a lifted representation for an \ac{fg} that possibly contains unknown factors.
We present the results of our empirical evaluation in \cref{sec:eval} before we conclude in \cref{sec:conclusion}.

\section{Preliminaries} \label{sec:prelim}
In this section, we begin by defining \acp{fg} as a propositional representation for a joint probability distribution between \acp{rv} and then introduce \acp{pfg}, which combine probabilistic models and first-order logic.
Thereafter, we describe the well-known \ac{cp} algorithm to lift a propositional model, i.e., to transform an \ac{fg} into a \ac{pfg} with equivalent semantics.

\subsection{Factor Graphs and Parameterised Factor Graphs}
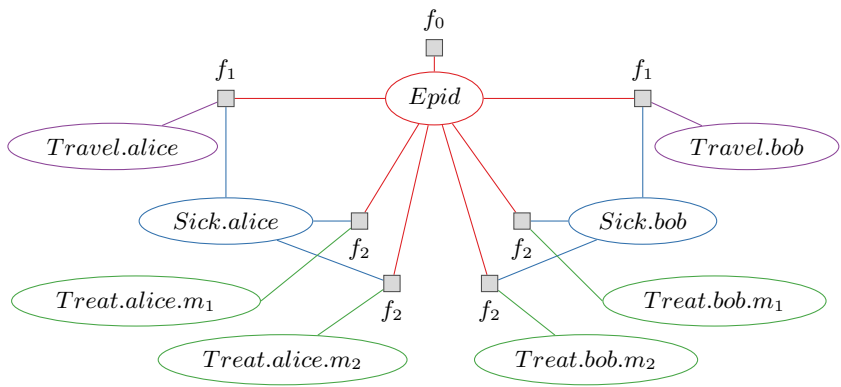
\begin{figure}[t]
	\centering
	\begin{tikzpicture}[
	rv/.append style={fill=white}
]
	\node[rv, draw=newred] (E) {$Epid$};

	\factor{above}{E}{0.2cm}{90}{$f_0$}{F0}

	\factor{left}{E}{2cm}{90}{$f_1$}{F1_1}
	\factor{right}{E}{2cm}{90}{$f_1$}{F1_2}

	\node[rv, draw=newblue, below = 1.2cm of F1_1] (SickA) {$Sick.alice$};
	\node[rv, draw=newblue, below = 1.2cm of F1_2] (SickB) {$Sick.bob$};

	\factor{right}{SickA}{0.5cm}{270}{$f_2$}{F2_1}
	\factor{below right}{F2_1}{0.6cm and 0.2cm}{270}{$f_2$}{F2_2}

	\factor{left}{SickB}{0.5cm}{270}{$f_2$}{F2_3}
	\factor{below left}{F2_3}{0.6cm and 0.2cm}{270}{$f_2$}{F2_4}

	\node[rv, draw=newpurple, below left = 0.3cm and 0.4cm of F1_1] (TravelA) {$Travel.alice$};
	\node[rv, draw=newpurple, below right = 0.3cm and 0.4cm of F1_2] (TravelB) {$Travel.bob$};
	\node[rv, draw=newgreen, below left = 0.6cm and -0.8cm of SickA] (TreatAM1) {$Treat.alice.m_1$};
	\node[rv, draw=newgreen, below right = 0.3cm and -0.4cm of TreatAM1] (TreatAM2) {$Treat.alice.m_2$};
	\node[rv, draw=newgreen, below right = 0.6cm and -0.8cm of SickB] (TreatBM1) {$Treat.bob.m_1$};
	\node[rv, draw=newgreen, below left = 0.3cm and -0.4cm of TreatBM1] (TreatBM2) {$Treat.bob.m_2$};

	\begin{pgfonlayer}{bg}
		\draw[newred] (E) -- (F0);
		\draw[newred] (E) -- (F1_1);
		\draw[newred] (E) -- (F2_1);
		\draw[newred] (E) -- (F2_2);
		\draw[newred] (E) -- (F1_2);
		\draw[newred] (E) -- (F2_3);
		\draw[newred] (E) -- (F2_4);
		\draw[newblue] (SickA) -- (F1_1);
		\draw[newblue] (SickA) -- (F2_1);
		\draw[newblue] (SickA) -- (F2_2);
		\draw[newpurple] (TravelA) -- (F1_1);
		\draw[newgreen] (TreatAM1.east) -- (F2_1);
		\draw[newgreen] (TreatAM2) -- (F2_2);
		\draw[newblue] (SickB) -- (F1_2);
		\draw[newblue] (SickB) -- (F2_3);
		\draw[newblue] (SickB) -- (F2_4);
		\draw[newpurple] (TravelB) -- (F1_2);
		\draw[newgreen] (TreatBM1.west) -- (F2_3);
		\draw[newgreen] (TreatBM2) -- (F2_4);
	\end{pgfonlayer}
\end{tikzpicture}
	\caption{An \ac{fg} for an epidemic example~\cite{Hoffmann2022a} with two individuals $alice$ and $bob$. The input-output pairs of the factors are omitted for simplification.}
	\label{fig:epid_fg}
\end{figure}

An \ac{fg} is an undirected graphical model to represent a full joint probability distribution between \acp{rv}~\cite{Kschischang2001a}.
In particular, an \ac{fg} is a bipartite graph that consists of two disjoint sets of nodes (variable nodes and factor nodes) with edges between a variable node $R$ and a factor node $f$ if the factor $f$ depends on $R$.
A factor is a function that maps its arguments to a positive real number (called potential).
The semantics of an \ac{fg} is given by $P(R_1, \dots, R_n) = \frac{1}{Z} \prod_f f$ with $Z$ being the normalisation constant.
\Cref{fig:epid_fg} shows an \ac{fg} representing an epidemic example with two individuals ($alice$ and $bob$) as well as two possible medications ($m_1$ and $m_2$) for treatment.
For each individual, there are two Boolean \acp{rv} $Sick$ and $Travel$, indicating whether the individual is sick and travels, respectively.
Moreover, there is another Boolean \ac{rv} $Treat$ for each combination of individual and medication, specifying whether the individual is treated with the medication.
The Boolean \ac{rv} $Epid$ states whether an epidemic is present.
Although the labelling of the nodes may suggest so, there is no explicit representation of individuals in the graph structure of the propositional \ac{fg}.
The names of the nodes only serve for the reader's understanding.

Clearly, the size of the \ac{fg} increases with an increasing number of individuals even though it is not necessary to distinguish between individuals because there are symmetries in the model (the factor $f_1$ occurs two times and the factor $f_2$ occurs four times).
In other words, the probability of an epidemic does not depend on knowing which specific individuals are being sick, but only on how many individuals are being sick.
To exploit such symmetries in a model, \acp{pfg} can be used.
We define \acp{pfg}, first introduced by Poole~\cite{Poole2003a}, based on the definitions given by Gehrke et al.~\cite{Gehrke2020a}.
\Acp{pfg} combine first-order logic with probabilistic models, using \acp{lv} as parameters in \acp{rv} to represent sets of indistinguishable \acp{rv}, forming \acp{prv}.

\begin{definition}[Logvar, PRV, Event]
	Let $\mathbf{R}$ be a set of \ac{rv} names, $\mathbf{L}$ a set of \ac{lv} names, $\Phi$ a set of factor names, and $\mathbf{D}$ a set of constants.
	All sets are finite.
	Each \ac{lv} $L$ has a domain $\mathcal{D}(L) \subseteq \mathbf{D}$.
	A \emph{constraint} is a tuple $(\mathcal{X}, C_{\mathcal{X}})$ of a sequence of \acp{lv} $\mathcal{X} = (X^1, \dots, X^n)$ and a set $C_{\mathcal{X}} \subseteq \times_{i = 1}^n\mathcal{D}(X_i)$.
	The symbol $\top$ for $C$ marks that no restrictions apply, i.e., $C_{\mathcal{X}} = \times_{i = 1}^n\mathcal{D}(X_i)$.
	A \emph{\ac{prv}} $R(L_1, \dots, L_n)$, $n \geq 0$, is a syntactical construct of a \ac{rv} $R \in \mathbf{R}$ possibly combined with \acp{lv} $L_1, \dots, L_n \in \mathbf{L}$ to represent a set of \acp{rv}.
	If $n = 0$, the \ac{prv} is parameterless and forms a propositional \ac{rv}.
	A \ac{prv} $A$ (or \ac{lv} $L$) under constraint $C$ is given by $A_{|C}$ ($L_{|C}$). 
	We may omit $|\top$ in $A_{|\top}$ or $L_{|\top}$.
	The term $\mathcal{R}(A)$ denotes the possible values (range) of a \ac{prv} $A$. 
	An \emph{event} $A = a$ denotes the occurrence of \ac{prv} $A$ with range value $a \in \mathcal{R}(A)$ and we call a set of events $\mathbf E = \{A_1 = a_1, \dots, A_k = a_k\}$ \emph{evidence}.
\end{definition}
As an example, consider $\mathbf{R} = \{Epid, Travel, Sick, Treat\}$ and $\mathbf{L} = \{X,M\}$ with $\mathcal{D}(X) = \{alice, bob\}$ (people), $\mathcal{D}(M) = \{m_1, m_2\}$ (medications), combined into Boolean \acp{prv} $Epid$, $Travel(X)$, $Sick(X)$, and $Treat(X,M)$.

A \ac{pf} describes a function, mapping argument values to positive real numbers (potentials), of which at least one is non-zero.

\begin{figure}[t]
	\centering
	\begin{tikzpicture}[
]
	\node[rv, draw=newred] (E) {$Epid$};
	\node[rv, draw=newblue, below = 0.8cm of E] (S) {$Sick(X)$};
	\node[rv, draw=newpurple, left = 0.3cm of S] (Travel) {$Travel(X)$};
	\node[rv, draw=newgreen, right = 0.3cm of S] (Treat) {$Treat(X,M)$};
	\factor{above}{E}{0.2cm}{90}{$g_0$}{G0}
	\pfs{above}{Travel}{0.5cm}{90}{$g_1$}{G1a}{G1}{G1b}
	\pfs{above}{Treat}{0.5cm}{90}{$g_2$}{G2a}{G2}{G2b}

	\begin{pgfonlayer}{bg}
		\draw[newred] (E) -- (G0);
		\draw[newred] (E) -- (G1);
		\draw[newred] (E) -- (G2);
		\draw[newblue] (S) -- (G1);
		\draw[newblue] (S) -- (G2);
		\draw[newpurple] (Travel) -- (G1);
		\draw[newgreen] (Treat) -- (G2);
	\end{pgfonlayer}
\end{tikzpicture}
	\caption{A \ac{pfg} corresponding to the lifted representation of the \ac{fg} depicted in \cref{fig:epid_fg}.The input-output pairs of the parfactors are again omitted for brevity.}
	\label{fig:epid_pm}
\end{figure}
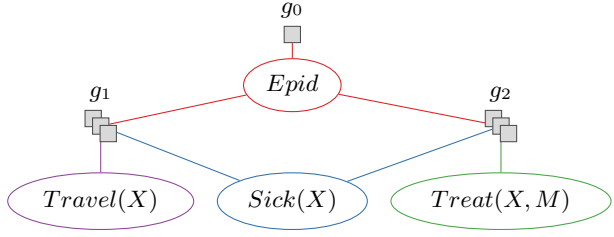

\begin{definition}[Parfactor, Model, Semantics]\label{def:pm}
	We denote a \emph{\ac{pf}} $g$ by $\phi(\mathcal{A})_{| C}$ with $\mathcal{A} = (A_1, \dots, A_n)$ a sequence of \acp{prv}, $\phi : \times_{i = 1}^n \mathcal{R}(A_i) \mapsto \mathbb{R}^+$ a function with name $\phi \in \Phi$, and $C$ a constraint on the \acp{lv} of $\mathcal{A}$. 
	We may omit $|\top$ in $\phi(\mathcal{A})_{| \top}$.
	The term $lv(Y)$ refers to the \acp{lv} in some element $Y$, a \ac{prv}, a \ac{pf}, or sets thereof.
	The term $gr(Y_{| C})$ denotes the set of all instances of $Y$ w.r.t.\ constraint $C$.
	A set of \acp{pf} $\{g_i\}_{i=1}^n$ forms a \emph{\ac{pfg}} $G$.
	The semantics of $G$ is given by grounding and building a full joint distribution.
	With $Z$ as the normalisation constant, $G$ represents $P_G = \frac{1}{Z} \prod_{f \in gr(G)} f$.
\end{definition}
For example, \cref{fig:epid_pm} shows a \ac{pfg} $G = \{g_i\}^2_{i=0}$ with $g_0 = \phi_0(Epid)_{| \top}$, $g_1 = \phi_1(Travel(X),\allowbreak Sick(X),\allowbreak Epid)_{| \top}$, and $g_2 = \phi_2(Treat(X,M),\allowbreak Sick(X),\allowbreak Epid)_{| \top}$.
The \ac{pfg} illustrated in \cref{fig:epid_pm} is a lifted representation of the \ac{fg} shown in \cref{fig:epid_fg}.
Note that the definition of \acp{pfg} also includes \acp{fg}, as every \ac{fg} is a \ac{pfg} containing only parameterless \acp{rv}.

\subsection{The Colour Passing Algorithm}
The \ac{cp} algorithm~\cite{Ahmadi2013a,Kersting2009a} constructs a lifted representation for an \ac{fg} where all factors are known.
As \ac{lifg} generalises \ac{cp}, we briefly recap how the \ac{cp} algorithm works.
The idea is to find symmetries in an \ac{fg} based on potentials of factors, ranges and evidence of \acp{rv}, as well as on the graph structure.
Each \ac{rv} is assigned a colour depending on its range and evidence, meaning that \acp{rv} with identical ranges and identical evidence are assigned the same colour, and each factor is assigned a colour depending on its potentials, i.e., factors with the same potentials get the same colour.
The colours are then passed from every \ac{rv} to its neighbouring factors and vice versa.
Passing colours around is repeated until the groupings of identical colours do not change anymore.
In the end, \acp{rv} and factors, respectively, are grouped together based on their colour signatures.

\begin{figure}[t]
	\centering
	\begin{tikzpicture}[label distance=1mm]
	\node[circle, draw] (A) {$A$};
	\node[circle, draw] (B) [below = 0.5cm of A] {$B$};
	\node[circle, draw] (C) [below = 0.5cm of B] {$C$};
	\factor{below right}{A}{0.25cm and 0.5cm}{270}{$\phi_1$}{f1}
	\factor{below right}{B}{0.25cm and 0.5cm}{270}{$\phi_2$}{f2}

	\nodecolorshift{newgreen}{A}{Acol}{-2.1mm}{1mm}
	\nodecolorshift{newgreen}{B}{Bcol}{-2.1mm}{1mm}
	\nodecolorshift{newgreen}{C}{Ccol}{-2.1mm}{1mm}

	\factorcolor{newpurple}{f1}{f1col}
	\factorcolor{newpurple}{f2}{f2col}

	\draw (A) -- (f1);
	\draw (B) -- (f1);
	\draw (B) -- (f2);
	\draw (C) -- (f2);

	\node[circle, draw, right = 1.3cm of A] (A1) {$A$};
	\node[circle, draw, below = 0.5cm of A1] (B1) {$B$};
	\node[circle, draw, below = 0.5cm of B1] (C1) {$C$};
	\factor{below right}{A1}{0.25cm and 0.5cm}{270}{$\phi_1$}{f1_1}
	\factor{below right}{B1}{0.25cm and 0.5cm}{270}{$\phi_2$}{f2_1}

	\nodecolorshift{newgreen}{A1}{A1col}{-2.1mm}{1mm}
	\nodecolorshift{newgreen}{B1}{B1col}{-2.1mm}{1mm}
	\nodecolorshift{newgreen}{C1}{C1col}{-2.1mm}{1mm}

	\factorcolor{newgreen}{f1_1}{f1_1col1}
	\factorcolorshift{newgreen}{f1_1}{f1_1col2}{2.1mm}{0mm}
	\factorcolorshift{newpurple}{f1_1}{f1_1col3}{4.2mm}{0mm}
	\factorcolor{newgreen}{f2_1}{f2_1col1}
	\factorcolorshift{newgreen}{f2_1}{f2_1col2}{2.1mm}{0mm}
	\factorcolorshift{newpurple}{f2_1}{f2_1col3}{4.2mm}{0mm}

	\coordinate[right=0.1cm of A1, yshift=-0.1cm] (CA1);
	\coordinate[above=0.2cm of f1_1, yshift=-0.1cm] (Cf1_1);
	\coordinate[right=0.1cm of B1, yshift=0.12cm] (CB1);
	\coordinate[right=0.1cm of B1, yshift=-0.1cm] (CB1_1);
	\coordinate[below=0.2cm of f1_1, yshift=0.15cm] (Cf1_1b);
	\coordinate[above=0.2cm of f2_1, yshift=-0.1cm] (Cf2_1);
	\coordinate[right=0.1cm of C1, yshift=0.12cm] (CC1);
	\coordinate[below=0.2cm of f2_1, yshift=0.15cm] (Cf2_1b);

	\begin{pgfonlayer}{bg}
		\draw (A1) -- (f1_1);
		\draw [arc, gray] (CA1) -- (Cf1_1);
		\draw (B1) -- (f1_1);
		\draw [arc, gray] (CB1) -- (Cf1_1b);
		\draw (B1) -- (f2_1);
		\draw [arc, gray] (CB1_1) -- (Cf2_1);
		\draw (C1) -- (f2_1);
		\draw [arc, gray] (CC1) -- (Cf2_1b);
	\end{pgfonlayer}

	\node[circle, draw, right = 1.6cm of A1] (A2) {$A$};
	\node[circle, draw, below = 0.5cm of A2] (B2) {$B$};
	\node[circle, draw, below = 0.5cm of B2] (C2) {$C$};
	\factor{below right}{A2}{0.25cm and 0.5cm}{270}{$\phi_1$}{f1_2}
	\factor{below right}{B2}{0.25cm and 0.5cm}{270}{$\phi_2$}{f2_2}

	\nodecolorshift{newgreen}{A2}{A2col}{-2.1mm}{1mm}
	\nodecolorshift{newgreen}{B2}{B2col}{-2.1mm}{1mm}
	\nodecolorshift{newgreen}{C2}{C2col}{-2.1mm}{1mm}

	\factorcolor{newpurple}{f1_2}{f1_2col1}
	\factorcolor{newpurple}{f2_2}{f2_2col1}

	\draw (A2) -- (f1_2);
	\draw (B2) -- (f1_2);
	\draw (B2) -- (f2_2);
	\draw (C2) -- (f2_2);

	\node[circle, draw, right = 1.3cm of A2] (A3) {$A$};
	\node[circle, draw, below = 0.5cm of A3] (B3) {$B$};
	\node[circle, draw, below = 0.5cm of B3] (C3) {$C$};
	\factor{below right}{A3}{0.25cm and 0.5cm}{270}{$\phi_1$}{f1_3}
	\factor{below right}{B3}{0.25cm and 0.5cm}{270}{$\phi_2$}{f2_3}

	\nodecolorshift{newpurple}{A3}{A3col1}{-4.2mm}{1mm}
	\nodecolorshift{newgreen}{A3}{A3col2}{-2.1mm}{1mm}
	\nodecolorshift{newpurple}{B3}{B3col1}{-6.3mm}{1mm}
	\nodecolorshift{newpurple}{B3}{B3col2}{-4.2mm}{1mm}
	\nodecolorshift{newgreen}{B3}{B3col3}{-2.1mm}{1mm}
	\nodecolorshift{newpurple}{C3}{C3col1}{-4.2mm}{1mm}
	\nodecolorshift{newgreen}{C3}{C3col2}{-2.1mm}{1mm}

	\factorcolor{newpurple}{f1_3}{f1_3col1}
	\factorcolor{newpurple}{f2_3}{f2_3col1}

	\coordinate[right=0.1cm of A3, yshift=-0.1cm] (CA3);
	\coordinate[above=0.2cm of f1_3, yshift=-0.1cm] (Cf1_3);
	\coordinate[right=0.1cm of B3, yshift=0.12cm] (CB3);
	\coordinate[right=0.1cm of B3, yshift=-0.1cm] (CB1_3);
	\coordinate[below=0.2cm of f1_3, yshift=0.15cm] (Cf1_3b);
	\coordinate[above=0.2cm of f2_3, yshift=-0.1cm] (Cf2_3);
	\coordinate[right=0.1cm of C3, yshift=0.12cm] (CC3);
	\coordinate[below=0.2cm of f2_3, yshift=0.15cm] (Cf2_3b);

	\begin{pgfonlayer}{bg}
		\draw (A3) -- (f1_3);
		\draw [arc, gray] (Cf1_3) -- (CA3);
		\draw (B3) -- (f1_3);
		\draw [arc, gray] (Cf1_3b) -- (CB3);
		\draw (B3) -- (f2_3);
		\draw [arc, gray] (Cf2_3) -- (CB1_3);
		\draw (C3) -- (f2_3);
		\draw [arc, gray] (Cf2_3b) -- (CC3);
	\end{pgfonlayer}

	\node[circle, draw, right = 1.3cm of A3] (A4) {$A$};
	\node[circle, draw, below = 0.5cm of A4] (B4) {$B$};
	\node[circle, draw, below = 0.5cm of B4] (C4) {$C$};
	\factor{below right}{A4}{0.25cm and 0.5cm}{270}{$\phi_1$}{f1_4}
	\factor{below right}{B4}{0.25cm and 0.5cm}{270}{$\phi_2$}{f2_4}

	\nodecolorshift{newgreen}{A4}{A4col}{-2.1mm}{1mm}
	\nodecolorshift{newblue}{B4}{B4col}{-2.1mm}{1mm}
	\nodecolorshift{newgreen}{C4}{C4col}{-2.1mm}{1mm}

	\factorcolor{newpurple}{f1_4}{f1_4col1}
	\factorcolor{newpurple}{f2_4}{f2_4col1}

	\draw (A4) -- (f1_4);
	\draw (B4) -- (f1_4);
	\draw (B4) -- (f2_4);
	\draw (C4) -- (f2_4);

	\pfs{right}{B4}{2.8cm}{270}{$\phi'_1$}{f12a}{f12}{f12b}

	\node[ellipse, inner sep = 1.2pt, draw, above left = 0.25cm and 0.5cm of f12] (AC) {$R(X)$};
	\node[circle, draw] (B) [below left = 0.25cm and 0.7cm of f12] {$B$};

	\begin{pgfonlayer}{bg}
		\draw (AC) -- (f12);
		\draw (B) -- (f12);
	\end{pgfonlayer}
\end{tikzpicture}
	\caption{The colour passing procedure of the \ac{cp} algorithm on an exemplary input \ac{fg} containing three Boolean \acp{rv} without evidence and two factors with identical potentials. The example has been introduced by Ahmadi et al.~\cite{Ahmadi2013a}.}
	\label{fig:cp_example}
\end{figure}
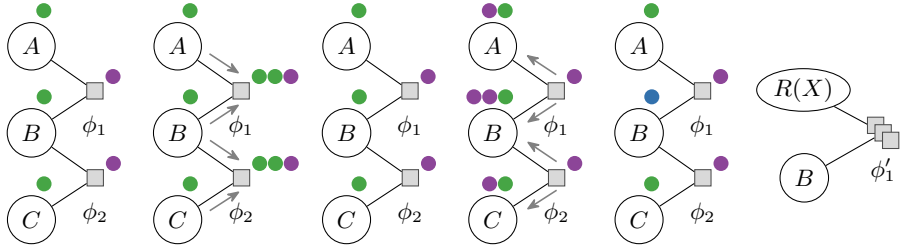

\Cref{fig:cp_example} depicts the procedure of the \ac{cp} algorithm on a simple \ac{fg}.
The two factors $\phi_1$ and $\phi_2$ share identical potentials in this example.
As all three \acp{rv} are Boolean and there is no evidence available, $A$, $B$, and $C$ are assigned the same colour (e.g., green).
Furthermore, the potentials of $\phi_1$ and $\phi_2$ are identical, so they are assigned the same colour (e.g., purple).
The colours are then passed from \acp{rv} to factors: $\phi_1$ receives two times the colour green from $A$ and $B$ and $\phi_2$ receives two times the colour green from $B$ and $C$.
Afterwards, $\phi_1$ and $\phi_2$ are recoloured according to the colours they received from their neighbours.
Since both $\phi_1$ and $\phi_2$ received the same colours, they are assigned the same colour during recolouring (e.g., purple).
The colours are then passed from factors to \acp{rv}.
During this step, not only the colours are shared but also the position of the \acp{rv} in the argument list of the corresponding factor.
Thus, $A$ receives a tuple $(\mathrm{purple}, 1)$ from $\phi_1$, $B$ receives $(\mathrm{purple}, 2)$ from $\phi_1$ and $(\mathrm{purple}, 2)$ from $\phi_2$, and $C$ receives $(\mathrm{purple}, 1)$ from $\phi_2$.
Building on these new colour signatures, the \acp{rv} are recoloured such that $A$ and $C$ receive the same colour while $B$ is assigned a different colour.
Iterating the colour passing procedure does not change these groupings and thus we obtain the \ac{pfg} shown on the right in \cref{fig:cp_example}.

When facing a situation with unknown factors being present in an \ac{fg}, the \ac{cp} algorithm cannot be applied to construct a lifted representation for the \ac{fg}.
In the upcoming section, we introduce the \ac{lifg} algorithm which generalises the \ac{cp} algorithm and is able to handle the presence unknown factors.

\section{The \acs{lifg} Algorithm} \label{sec:lifg}
As our goal is to perform lifted inference, we have to obtain a \ac{pfg} where all potentials are known.
To transform an \ac{fg} containing unknown factors into a \ac{pfg} without unknown factors, we transfer potentials from known factors to unknown factors.
For example, consider again the \ac{fg} depicted in \cref{fig:epid_fg} and assume that another individual, say $eve$, is added to the model.
Like $alice$ and $bob$, $eve$ can travel, be sick, and be treated and hence, four new \acp{rv} with three new corresponding factors are attached to the model.
However, as we might have limited data, we might not always know the exact potentials for the newly introduced factors when a new individual is added to the model and thus, we end up with a model containing unknown factors.
In this example, we can transfer the potentials of the known factors $f_1$ and $f_2$ to the newly introduced unknown factors, as it is reasonable to assume that $eve$ behaves the same as $alice$ and $bob$ as long as no evidence suggesting the contrary is available.

In an \ac{fg} containing unknown factors, the only information available to measure the similarity of factors is the neighbouring graph structure of the factors.
For the upcoming definitions, let $\Ne_G(v)$ denote the set of neighbours of a node $v$ (variable node or factor node) in $G$, i.e., $\Ne_G(f)$ contains all \acp{rv} connected to a factor $f$ in $G$ and $\Ne_G(R)$ contains all factors connected to a \ac{rv} $R$ in $G$.
If the context is clear, we omit the subscript from $\Ne_G(v)$ and write $\Ne(v)$ for simplification.
We start by defining the 2-step neighbourhood of a factor $f$ as the set containing all \acp{rv} that are connected to $f$ as well as all factors connected to a \ac{rv} that is connected to $f$.
The concept of taking into account all nodes with a maximal distance of two is based on the idea of a single iteration of the colour passing procedure.

\begin{definition}[2-Step Neighbourhood]
	The \emph{2-step neighbourhood} of a factor $f$ in an \ac{fg} $G$ is defined as
	\begin{align*}
		\twostepG{G}{f} = \{ R \mid R \in \Ne_G(f) \} \cup \{ f' \mid \exists R: R \in \Ne_G(f) \land f' \in \Ne_G(R) \}.
	\end{align*}
\end{definition}
If the context is clear, we write \twostep{f} instead of \twostepG{G}{f}.
For example, the 2-step neighbourhood of $\phi_1$ in the \ac{fg} depicted in \cref{fig:cp_example} is given by $\twostep{\phi_1} = \{A,B\} \cup \{\phi_1, \phi_2\}$.
By $G[V']$ we denote the subgraph of a graph $G$ induced by a subset of nodes $V'$, that is, $G[V']$ contains only the nodes in $V'$ as well as all edges from $G$ that connect two nodes in $V'$.
In our example, $G[\twostep{\phi_1}]$ then consists of the nodes $A$, $B$, $\phi_1$, and $\phi_2$, and contains the edges $A - \phi_1$, $B - \phi_1$, and $B - \phi_2$.
As it is currently unknown whether a general graph isomorphism test is solvable in polynomial time, we make use of the notion of symmetric 2-step neighbourhoods instead of relying on isomorphic 2-step neighbourhoods to ensure that \ac{lifg} is implementable in polynomial time.

\begin{definition}[Symmetric 2-Step Neighbourhoods]
	Given an \ac{fg} $G$ and factors $f_i$, $f_j$ in $G$, $G[\twostepG{G}{f_i}]$ is \emph{symmetric} to $G[\twostepG{G}{f_j}]$ if
	\begin{enumerate}
		\item $\abs{\Ne_G(f_i)} = \abs{\Ne_G(f_j)}$ and
		\item there exists a bijection $\phi: \Ne_G(f_i) \to \Ne_G(f_j)$ that maps every \ac{rv} $R_k \in \Ne_G(f_i)$ to a \ac{rv} $R_{\ell} \in \Ne_G(f_j)$ such that the evidence for $R_k$ and $R_{\ell}$ is identical, $\mathcal R(R_k) = \mathcal R(R_{\ell})$, and $\abs{\Ne_G(R_k)} = \abs{\Ne_G(R_{\ell})}$.
	\end{enumerate}
\end{definition}
For example, take a look again at the \ac{fg} shown in \cref{fig:cp_example} and assume that there is no evidence.
We can check whether $\phi_1$ and $\phi_2$ have symmetric 2-step neighbourhoods: Both $\phi_1$ and $\phi_2$ are connected to two \acp{rv} as $\Ne(\phi_1) = \{A,B\}$ and $\Ne(\phi_2) = \{B,C\}$, thereby satisfying the first condition.
Further, $A$ can be mapped to $C$ with $\mathcal R(A) = \mathcal R(C)$ (Boolean) and $\abs{\Ne(A)} = \abs{\Ne(C)} = 1$ and $B$ can be mapped to itself.
Thus, condition two is satisfied and it holds that $G[\twostep{\phi_1}]$ is symmetric to $G[\twostep{\phi_2}]$.
Having defined the notion of symmetric 2-step neighbourhoods, we are able to specify a condition for two factors to be possibly identical.
Two factors are considered possibly identical if the subgraphs induced by their 2-step neighbourhoods are symmetric.

\begin{definition}[Possibly Identical Factors]
	Given two factors $f_i$ and $f_j$ in an \ac{fg} $G$, we call $f_i$ and $f_j$ \emph{possibly identical}, denoted as $f_i \approx f_j$, if
	\begin{enumerate}
		\item $G[\twostepG{G}{f_i}]$ is symmetric to $G[\twostepG{G}{f_j}]$ and
		\item at least one of $f_i$ and $f_j$ is unknown, or $f_i$ and $f_j$ have the same potentials.
	\end{enumerate}
\end{definition}
The second condition serves to ensure consistency as two factors with different potentials can obviously not be identical.
Applying the definition of possibly identical factors to $\phi_1$ and $\phi_2$ from \cref{fig:cp_example}, we can verify that $\phi_1$ and $\phi_2$ are indeed possibly identical because they have symmetric 2-step neighbourhoods and identical potentials.
Next, we describe the entire \ac{lifg} algorithm, which is illustrated in \cref{alg:lifg}.

\begin{algorithm}[t]
	\SetKwInOut{Input}{Input}
	\SetKwInOut{Output}{Output}
	\caption{\acs{lifg}}
	\label{alg:lifg}
	\Input{An \ac{fg} $G$ with \acp{rv} $\mathbf{R} = \{R_1, \dots, R_n\}$, known factors $\mathbf{F} = \{f_1, \dots, f_m\}$, unknown factors $\mathbf{F'} = \{f'_1, \dots, f'_z\}$, and evidence $\mathbf{E} = \{R_1 = r_1, \dots, R_k = r_k\}$, and a real-valued threshold $\theta \in [0, 1]$.}
	\Output{A lifted representation $G'$ of $G$.}
	\BlankLine
	Assign each $f_i \in \mathbf{F}$ a colour based on its potentials\;
	Assign each $f'_i \in \mathbf{F'}$ a unique colour\;
	\ForEach{unknown factor $f_i \in \mathbf{F'}$}{
		$C_{f_i} \gets \{\}$\;
		\ForEach{factor $f_j \in \mathbf{F} \cup \mathbf{F'}$ with $f_i \neq f_j$}{
			\If{$f_i \approx f_j$}{
				\eIf{$f_j$ is unknown}{
					Assign $f_j$ the same colour as $f_i$\;
				}{
					$C_{f_i} \gets C_{f_i} \cup \{f_j\}$\;
				}
			}
		}
	}
	\ForEach{set of candidates $C_{f_i}$}{
		$C_{f_i}^{\ell} \gets$ Maximal subset of $C_{f_i}$ such that $f_j \approx f_k$ holds for all $f_j, f_k \in C_{f_i}^{\ell}$\;
		\If{$\abs{C_{f_i}^{\ell}} \mathbin{/} \abs{C_{f_i}} \geq \theta$}{
			Assign all $f_j \in C_{f_i}^{\ell}$ the same colour as $f_i$\;
		}
	}
	$G \gets$ Result from calling the \ac{cp} algorithm on the modified graph $G$ and $\mathbf{E}$\;
\end{algorithm}

\Ac{lifg} assigns colours to unknown factors based on symmetric subgraphs induced by their 2-step neighbourhoods, proceeding as follows for an input $G$.
As an initialisation step, \ac{lifg} assigns each known factor a colour based on its potentials and each unknown factor a unique colour.
Then, \ac{lifg} searches for possibly identical factors in two phases.
In the first phase, all unknown factors that are possibly identical are assigned the same colour, as there is no way to distinguish them.
Furthermore, \ac{lifg} collects for every unknown factor $f_i$ a set $C_{f_i}$ of known factors possibly identical to $f_i$.
The second phase then continues to group the unknown factors with known factors, including the transfer of the potentials from the known factors to the unknown factors.
For every unknown factor $f_i$, \ac{lifg} computes a maximal subset $C_{f_i}^{\ell} \subseteq C_{f_i}$ for which all elements are pairwise possibly identical.
Afterwards, $f_i$ and all $f_j \in C_{f_i}^{\ell}$ are assigned the same colour if a user-defined threshold is reached.
Finally, \ac{cp} is called on $G$, which now includes the previously set colours for the unknown factors in $G$, to group both known and unknown factors in $G$.

The purpose of the threshold $\theta$ is to control the required agreement of known factors before grouping unknown factors with known factors as it is possible for an unknown factor to be possibly identical to multiple known factors having different potentials.
A larger $\theta$ requires a higher agreement, e.g., $\theta = 1$ requires all candidates to have identical potentials.
Note that all known factors in $C_{f_i}^{\ell}$ are guaranteed to have identical potentials (otherwise they would not be pairwise possibly identical) and thus, their potentials can be transferred to $f_i$.
Consequently, the output of \ac{lifg} is guaranteed to contain only known factors and hence ensures a well-defined semantics if $C_{f_i}^{\ell}$ is non-empty for each unknown factor $f_i$ and the threshold is sufficiently small (e.g., zero) to group each unknown factor with at least one known factor.
\begin{corollary}
	Given that for every unknown factor $f_i$ there is at least one known factor that is possibly identical to $f_i$ in an \ac{fg} $G$, \ac{lifg} is able to replace all unknown potentials in $G$ by known potentials.
\end{corollary}
It is easy to see that \ac{lifg} is a generalisation of \ac{cp}, meaning that both algorithms compute the same result for input \acp{fg} containing only known factors (if an input \ac{fg} $G$ contains no unknown factors, only the first line and the last line of \cref{alg:lifg} are executed---which is equivalent to calling \ac{cp} on $G$).
\begin{corollary}
	Given an \ac{fg} that contains only known factors, \ac{cp} and \ac{lifg} output identical groupings of \acp{rv} and factors, respectively.
\end{corollary}
Next, we investigate the practical performance of \ac{lifg} in our evaluation.

\section{Empirical Evaluation} \label{sec:eval}
In this section, we present the results of the empirical evaluation for \ac{lifg}.
To evaluate the performance of \ac{lifg}, we start with a non-parameterised \ac{fg} $G$ where all factors are known, serving as our ground truth.
Afterwards, we remove the potential mappings for five to ten percent of the factors in $G$, yielding an incomplete \ac{fg} $G'$ on which \ac{lifg} is run to obtain a \ac{pfg} $G_{\textsc{\ac{lifg}}}$.
Each factor $f'$ whose potentials are removed is chosen randomly under the constraint that there exists at least one other factor with known potentials that is possibly identical to $f'$.
This constraint corresponds to the assumption that there exists at least one group to which a new individual can be added and it ensures that after running \ac{lifg}, probabilistic inference can be performed for evaluation purposes.
Clearly, in our evaluation setting, there is not only a single new individual but instead a set of new individuals, given by the set of factors whose potentials are missing.
We use a parameter $d=2,4,8,16,32,64,128,256$ to control the size of the \ac{fg} $G$ (and thus, the size of $G'$).
More precisely, for each choice of $d$, we evaluate multiple input \acp{fg} which contain between $2d$ and $3d$ \acp{rv} (and factors, respectively).
The potentials of the factors are randomly generated such that the ground truth $G$ contains between three and five (randomly chosen) cohorts of \acp{rv} which should be grouped together, with one cohort containing roughly 50 percent of all \acp{rv} in $G$ while the other cohorts share the remaining 50 percent of the \acp{rv} from $G$ uniformly at random.

We set $\theta = 0$ to ensure that each unknown factor is grouped with at least one known factor to be able to perform lifted probabilistic inference on $G_{\textsc{\ac{lifg}}}$ for evaluation.
To assess the error made by \ac{lifg} for each choice of $d$, we pose $d$ different queries to the ground truth $G$ and to $G_{\textsc{\ac{lifg}}}$, respectively.
For each query, we compute the \ac{kl} divergence~\cite{Kullback1951a} between the resulting probability distributions for the ground truth $G$ and $G_{\textsc{\ac{lifg}}}$ to measure the similarity of the query results.
The \ac{kl} divergence measures the difference of two distributions and its value is zero if the distributions are identical.

\begin{figure}[t]
	\centering
	\resizebox{\textwidth}{!}{\input{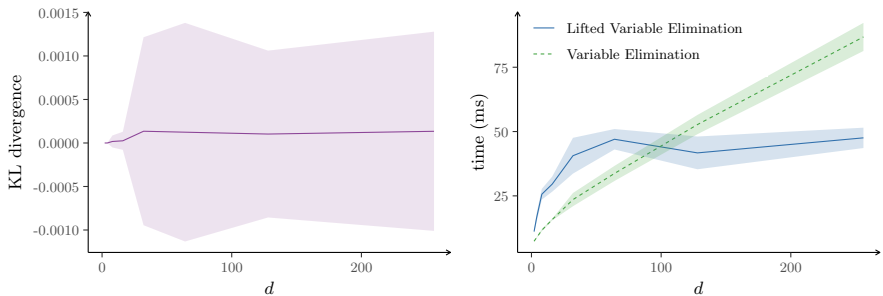}}
	\caption{Left: The mean \ac{kl} divergence on the queried probability distributions (thick line) as well as the standard deviation of all measured \ac{kl} divergences for each choice of $d$ (ribbon around the mean). Right: The mean run time of \acl{ve} and \acl{lve} for each choice of $d$.}
	\label{fig:plot-eval}
\end{figure}

In the left plot of \cref{fig:plot-eval}, we report the mean \ac{kl} divergence over all queries for each choice of $d$.
The ribbon around the line illustrates the standard deviation of the measured \ac{kl} divergences.
We find that the mean \ac{kl} divergence is close to zero for all choices of $d$ in practice.
Both the mean \ac{kl} divergence and the standard deviation of the \ac{kl} divergences do not show any significant differences between the various values for $d$.
Note that the depicted standard deviation is also very small for all choices of $d$ due to the granularity of the y-axis.
The maximum \ac{kl} divergence measured for any choice of $d$ is about $0.01$.

Given our assumptions, a new individual actually belongs to a cohort and most cohorts behave not completely different.
So normally, we trade off accuracy of query results for the ability to perform inference, which otherwise would not be possible at all.
If the semantics cannot be fixed, missing potentials need to be guessed to be able to perform inference at all, probably resulting in worse errors.
As we basically perform unsupervised clustering, errors might happen when grouping unknown factors with known factors.
The error might be further reduced by increasing the effort when searching for known factors that are possible candidates for grouping with an unknown factor.
For example, it is conceivable to increase the size of the neighbourhood during the search for possible identical factors at the expense of a higher run time expenditure.

In addition to the error measured by the \ac{kl} divergence, we also report the run times of \acl{ve} on $G$ and \acl{lve} on the \ac{pfg} computed by \ac{lifg}, i.e., $G_{\textsc{\ac{lifg}}}$.
The run times are shown in the right plot of \cref{fig:plot-eval}.
As expected, \acl{lve} is faster than \acl{ve} for larger graphs and the run time of \acl{lve} increases more slowly with increasing graph sizes than the run time of \acl{ve}.
Hence, \ac{lifg} not only allows to perform probabilistic inference at all, but also speeds up inference by allowing for lifting probabilistic inference.
Note that there are on average 24 different groups over all settings with the largest domain size being 87 (for the setting of $d=256$), i.e., there are a lot of small groups (of size one) which diminish the advantage of \acl{lve} over \acl{ve}.
We could also obtain more compact \acp{pfg} by merging groups that are not fully identical but similar to a given extent such that the resulting \ac{pfg} contains less different groups at the cost of a lower accuracy of query results.
Obtaining a more compact \ac{pfg} would most likely result in a higher speedup of \acl{lve} compared to \acl{ve}.

\section{Conclusion} \label{sec:conclusion}
In this paper, we introduce the \ac{lifg} algorithm to construct a lifted representation for an \ac{fg} that possibly contains unknown factors.
\Ac{lifg} is a generalisation of the widespread \ac{cp} algorithm and allows to transfer potentials from known factors to unknown factors by identifying symmetric subgraphs.
Under the assumption that for every unknown factor there exists at least one known factor having a symmetric surrounding graph structure to it, \ac{lifg} is able to replace all unknown potentials in an \ac{fg} by known potentials.

\section*{Acknowledgements}
This work was partially supported by the BMBF project AnoMed.
The research of Malte Luttermann was also partially supported by the Medical Cause and Effects Analysis (MCEA) project.
This preprint has not undergone peer review or any post-submission improvements or corrections.
The Version of Record of this contribution is published in \emph{Lecture Notes in Computer Science, Volume 14294}, and is available online at \url{https://doi.org/10.1007/978-3-031-45608-4_25}.

\bibliographystyle{splncs04}
\bibliography{references.bib}
\end{document}